\documentclass[twoside,11pt]{article}

\usepackage{jmlr2e}

\usepackage{color,amsmath,amssymb,mathtools} 
\usepackage{natbib}  
\usepackage{caption} 
\usepackage[switch]{lineno}
\usepackage[ruled,vlined]{algorithm2e}
\usepackage{subcaption,amsmath,amssymb}
\newcommand{\e}{epidemiological}
\newcommand{\epi}{EpiCast}

\SetKw{KwBy}{by}

\ShortHeadings{Machine Learning-Powered Mitigation Policy Optimization in Epidemiological Models}{Jayaraman J. Thiagarajan \textit{et al.}}
\firstpageno{1}

\begin{document}
	\title{Machine Learning-Powered Mitigation Policy Optimization in Epidemiological Models}
	\author{\name Jayaraman J. Thiagarajan \email jjayaram@llnl.gov\\
		\addr Lawrence Livermore National Laboratory \\
		Livermore, CA, USA
		\AND
		\name Peer-Timo Bremer \email bremer5@llnl.gov \\
		\addr Lawrence Livermore National Laboratory \\
		Livermore, CA, USA
	\AND
	\name Rushil Anirudh \email anirudh1@llnl.gov \\
	\addr Lawrence Livermore National Laboratory \\
	Livermore, CA, USA
\AND
\name Timothy C. Germann  \email tcg@lanl.gov \\
\addr Los Alamos National Laboratory \\
Los Alamos, NM, USA
\AND
\name Sara Y. Del Valle \email sdelvall@lanl.gov \\
\addr Los Alamos National Laboratory \\
Los Alamos, NM, USA
\AND
\name Frederick H. Streitz  \email streitz1@llnl.gov \\
\addr  Lawrence Livermore National Laboratory \\
Livermore, CA, USA
}

\maketitle

\vspace{0.2in}

\begin{abstract}
A crucial aspect of managing a public health crisis is to effectively balance prevention and mitigation strategies, while taking their socio-economic impact into account. In particular, determining the influence of different non-pharmaceutical interventions (NPIs) on the effective use of public resources is an important problem, given the uncertainties on when a vaccine will be made available. In this paper, we propose a new approach for obtaining optimal policy recommendations based on \e~models, which can characterize the disease progression under different interventions, and a look-ahead reward optimization strategy to choose the suitable NPI at different stages of an epidemic. Given the time delay inherent in any \e~model and the exponential nature especially of an unmanaged epidemic, we find that such a look-ahead strategy infers non-trivial policies that adhere well to the constraints specified. Using two different \e~models, namely SEIR and EpiCast, we evaluate the proposed algorithm to determine the optimal NPI policy, under a constraint on the number of daily new cases and the primary reward being the absence of restrictions.
\end{abstract}

\section{Introduction}
\label{sec:intro}
One of the key challenges in managing a public health crisis is the allocation of scarce resources and how to balance the cost-benefits of mitigation strategies. This is especially true for non-pharmaceutical interventions (NPIs), which potentially impact a large number of otherwise not affected populations~\cite{ferguson2020report,morato2020optimal}. In order to asses the impact of optimal NPIs, one can parameterize epidemiological models to represent disease progression at different levels of interventions, and measure the predicted number of new infections~\cite{ghamizi2020data}. Conceptually, this can be used to formulate an optimization problem in the context of non-linear control, wherein users can assign a cost/reward to different levels of interventions and specify constraints  on the peak response~\cite{alvarez2020simple,yaesoubi2016identifying}. However, \e~models pose a number of unique challenges in both formulating the problem as well as solving it.

The first challenge is to find a meaningful formulation of what can be considered optimal under which potential constraints~\cite{khadilkar2020optimising, libin2020deep}.
An intuitive objective would be to minimize harm, i.e. aim to minimize the number of infected or potential casualties. However, this directly leads to a trivial solution of applying maximal NPIs (e.g. lockdown) at all times, which by design will lead to the fewest number of infections. However, this disregards any socio-economic costs, i.e. lost business activity or negative health effects due to inactivity. In practice, balancing the different aspects explicitly is challenging and requires not only in-depth economic studies~\cite{guerrieri2020macroeconomic} but also knowledge of when vaccines might become available or what role herd immunity will play as an exit strategy~\cite{lurie2020developing}, etc.
Instead, we use a common simplification which reformulates the problem in terms of health care management with the primary constraint being the number of daily new cases, and a reward assignment process that encourages lessened restrictions.
This reflects the desire to guarantee adequate care for all sick given the finite healthcare resources, and assumes that in isolation applying any NPI will incur a cost (monetary or otherwise) to be avoided.
As will be discussed in more detail below, our approach allows policy makers to assign relative rewards to different NPIs reflecting the differences between, for example, closing all non-essential businesses vs.\ preventing large gatherings.

The second challenge is the time delay inherent in any epidemiological model and the exponential nature of an unmanaged epidemic~\cite{liu2020microscopic,Germann2019}.
Due to effects such as incubation times or asymptomatic infections, interventions taken (or removed) today may show a significant effect only days or weeks later.
Simultaneously, allowing an exponential growth even within a given constraint may result in unavoidable future violations despite maximal NPIs.
Both of these issues emphasize the need for a substantial look-ahead window to consider not only the current state but a long term forecast.
Ideally, this window should be ``infinite'' to avoid any possibility of unexpected problems due to delayed dynamics.
However, this formulation makes the optimization expensive especially for more complex models.
Here, we present a greedy approach based on a two-level look-ahead strategy able to produce optimal policy recommendations for a variety of different scenarios.
Furthermore, the optimization incorporates practical considerations, such as, the limited frequencies at which public policies can be changed, or the variable cost of different interventions.
Finally, the two-level optimization provides an intuitive trade-off between short term gains and potential future costs beyond the total reward inherent in the system.
For example, the optimal policy might place relatively stringent early NPIs to avoid predicted peaks months later.
However, given the uncertainty of long term predictions one may prefer a more optimistic choice early as long as there is sufficient lead time to prevent peaks later even if this produces an overall lower reward. Using empirical studies with two popular \e~models, namely SEIR and agent-based EpiCast, we demonstrate the effectiveness of our approach.

\section{Problem Formulation}
We first provide a formal definition of the policy optimization problem. Without loss of generality, let us denote the current state of an epidemic, as described by a model $\mathrm{E}$, as $\mathrm{X}[t]$, where $t$ is the time step.
For example, $\mathrm{X}$ can refer to the set of susceptible, exposed, infected and removed populations in an SEIR model or a more complex internal state of an agent-based system.
One can evaluate the future states by evaluating the \e~model $\mathrm{X}[t+1:t+k] = \mathrm{E}(X[t]; \mathrm{n}[t],k)$, where $\mathrm{n}[t]$ denotes the NPI applied at time-step $t$.
In our formulation, we assume that each $n[t] \in \mathcal{N}$, where $\mathcal{N}$ is a set of discrete NPI choices ordered by severity, i.e. $\mathcal{N}(i)$ is more restrictive with a smaller reward than  $\mathcal{N}(i+1)$. 
Note that in practice, the NPI choices will typically be represented by some combination of model parameters depending on the model. 
For example, in case of an SEIR model, $\mathcal{N}$ represents a set of infection rates $\beta$ which correspond to more or less stringent NPIs.
We refer to a sequence of NPIs, $\mathcal{P} = \{n[1], n[2],\cdots,n[T]\}$, as a \textit{policy}.
With $\mathcal{C} \in \mathbb{R}^{|\mathcal{N}|}$ denoting the reward for adopting each of the NPI choices in $\mathcal{N}$, one can compute the reward for the policy as $\mathcal{R} = r[1] + r[2] + \cdots + r[T]\}$, $\forall r[t] \in \mathcal{C}$. Here, the symbol $|.|$ is the cardinality of a set.
The design objective of not overloading the healthcare system is specified by a threshold on the maximum number of new daily cases, denoted by $\tau$.
Note that, other constraints could be used, i.e. the number of currently active cases, predicted number of patients requiring the ICU, etc.
Finally, we consider policies in sets of $d$ days both to reflect the fact that public policy cannot realistically be changed on a day to day basis as well as to make the optimization more tractable. 
The goal is to maximize $\mathcal{R}$ while not violating $\tau$ anytime during the course of the epidemic. While a few studies have been recently proposed in the literature to utilize reinforcement learning approaches with a tractable \e{} model to optimize for mitigation policies~\cite{libin2020deep, khadilkar2020optimising, ghamizi2020data, liu2020microscopic}, our goal is to deal with complex models, such as the agent-based EpiCast considered in our study, which is computationally expensive to be repeatedly evaluated. Furthermore, we are interested in designing a scalable optimization approach that can be rapidly executed for a wide-variety of constraints and specifications, which is known to be a bottleneck for episodic-training based reinforcement learing methods. 

\section{Epidemiological Models}
In this section, we briefly describe the two \e~models used in our study and provide suitable references for additional details.
\subsection{SEIR Model}
We consider a compartmental SEIR model from which one can obtain trajectories of the epidemic, given the current state and epidemic-specific parameters. An SEIR model divides the population into \textit{Susceptible}, \textit{Exposed}, \textit{Infected} and \textit{Recovered} compartments and can be described in terms of ordinary differential equations (ODEs)~\cite{he2020seir}. While  exposed refers to the latent infected but not yet infectious population, recovered contains the population that is no longer infectious (also referred as removed). While this has been popularly used to model influenza epidemics~\cite{mills2004transmissibility}, there have been several existing efforts that have utilized this model with great success in the case of the recent COVID-19 epidemic~\cite{lopez2020modified,roda2020difficult,yang2020modified}. Formally, an SEIR model is described as follows:
\begin{equation}
    \begin{aligned}
     \frac{dS}{dt} &= -\beta S(t) I(t) \\
    \frac{dE}{dt} &= \beta S(t) I(t) - \sigma E(t)\\
    \frac{dI}{dt} &= \sigma E(t) - \gamma I(t) \\
     \frac{dR}{dt} &= \gamma I(t).
\end{aligned}
    \label{eqn:seir}
\end{equation}
Here, the the rate of changes in the compartments are parameterized by infectious rate $\beta$, incubation rate $\sigma$ and the recovery rate $\gamma$. The severity of an epidemic is characterized by the \textit{basic reproduction number} that quantifies the number of secondary infections from an individual in an entirely susceptible population as
$R_0 = \frac{\beta}{\gamma}$. Following common parameter settings assumed by COVID-19 studies in different countries, we set $\gamma = 0.1$, $\sigma = 0.2$. Interestingly, even in this simple model, one can introduce the effects of different NPI choices through changes in the infectious rate $\beta$. While there are existing works that attempt to estimate the current value of $\beta$ using additional data sources (e.g. mobility), in order to better fit the observed trajectory~\cite{soures2020sirnet}, our focus is on choosing the optimal policy of intervention. In particular, we define the set of NPI choices through corresponding $\beta$ values $\mathcal{N}\coloneqq \{0.25,0.3,0.5,0.7,0.8,0.9\}$, where higher $\beta$ implies lower restrictions.  

\subsection{Agent-based EpiCast Model}
EpiCast is an individual-based model, with daily contacts between people in household, workplace, school, neighborhood, and community settings. The primary data source is U.S. Census demographics at the tract level (the $\sim 65,000$ tracts are subsets of the $\sim 3000$ counties, with typically a few thousand people in each tract), and Census tract-to-tract workerflow data (i.e., how many people live in tract A and work in tract B). This is used to construct a model population with tract-level age and household size demographics, and realistic daily workflow pattern, which captures most of the short-range mobility. In addition, occasional long-distance travel is possible. A 12-hour timestep is used, so (unless on travel) individuals spend the night-time at home and day-time at school or workplace, if they belong to one (and they are open).  Additional details are provided in the Supporting Information of~\cite{Germann2006}. In the original model~\cite{Germann2006,Halloran2008}, the individual age- and context-specific contact rates that account for the duration and closeness of interactions between pairs of individuals in different settings (home, school, workplace, neighborhood, community, etc.) were uniform across the US. In a recent school dismissal study~\cite{Germann2019}, different communities were allowed to close their schools at different times, depending upon the current local disease incidence. In adapting this model to COVID-19, these local policies have been extended to all community mitigation measures: school dismissal, workplace closure, shelter-in-place, and other social distancing.

\begin{algorithm*}[t]
\SetAlgoLined
\SetKwInOut{Input}{input}
\SetKwInOut{Output}{output}
\Input{Epidemiological model $\mathrm{E}$; \newline Initial state $X[0]$; look-ahead parameters $k$, $k_s$; threshold $\tau$; frequency of NPI change $d$; Time steps $T$; \newline Set of NPI choices $\mathcal{N}$; Set of rewards for each of the NPIs $\mathcal{C}$.}
\Output{Policy $\mathcal{P}$, Reward $\mathcal{R}$}

\For{$t\gets1$ \KwTo $T$ \KwBy $d$}{
\For{$i\gets1$ \KwTo $|\mathcal{N}|$}{
$\bar{\mathrm{X}}[1:k] = E(\mathrm{X}[t-1]; \mathcal{N}(i), k)$ \tcc*[l]{Run $\mathrm{E}$ for $k$ steps with $\mathcal{N}(i)$}
Using $\bar{\mathrm{X}}[1:k]$, obtain the counts of new infected cases at each time step $N_c[1:k]$;

\If{$\max(N_c[1:k]) > \tau$}{
net$[i] = 0$ \tcc*[l]{reward for $\mathcal{N}(i)$}
break
}

net[i] = $\mathcal{C}(i)$*k\tcc*[l]{Compute the reward for $k$ steps} 

\For{$j\gets i$ \KwTo $1$ \KwBy $-1$}{
$\tilde{\mathrm{X}}[1:k_s] = E(\bar{\mathrm{X}}[k]; \mathcal{N}(j), k_s)$ \tcc*[l]{Run $\mathrm{E}$ for $k_s$ steps with $\mathcal{N}(j)$}

Using $\tilde{\mathrm{X}}[1:k_s]$, obtain the counts of new infected cases at each time step $\tilde{N}_c[1:k_s]$;

\tcc{Find the first index where the threshold is violated}
\For{$ind \gets 1$ \KwTo $k_s$}{
\If{$\tilde{N}_c[ind] > \tau$}{
break
}
}
$f[j] = (ind-1)*\mathcal{C}(j)$ \tcc*[l]{long-term reward when one NPI switch is allowed}
}
net$[i] += \max(f)$
}
$n = \arg \max($net$)$\tcc*{use NPI with the highest net score for $d$ time steps}

$\mathcal{P}[t:t+d] =  \mathcal{N}(n)$;

$\mathcal{R}[t:t+d] = \mathcal{C}(n)$;

$\mathrm{X}[t:t+d] = E(\mathrm{X}[t-1]; \mathcal{N}(n), d)$\tcc*{Evaluate $\mathrm{E}$ for $d$ steps with the chosen NPI }

}

\caption{Mitigation policy optimization.}
\label{alg}
\end{algorithm*}



\section{Proposed Algorithm}
We now describe the new two-level optimization used to solve the problem described above, followed by a discussion on dealing with models that are computationally expensive to be optimized directly.

The key novelty of our approach for solving the above optimization problem is to split the reward using a two-level scheme with a finite time horizon that enables an efficient early termination of a greedy search. 
In particular, policy choices are made every $d$ days based on combining a short-term and a long-term reward:
Short-term awards are computed from the current day for $k$ days forward and a policy choice is given its full reward if the constraint is met for all $k$ days and no reward if it is predicted to violate the constraint at any point.
This is equivalent to a brute force search of all policy choices and the straight-forward check of the constraint, both restricted to a short-term forecast that can be computed efficiently.
Note that typically $k > d$ which represents a first restriction on the search space, as in principle a switch to more stringent NPIs at the next $d$-day interval may prevent the constraint violation.
However, for reasonable $k$ and given the delayed response of the disease progression it is likely that any NPI that would violate the constraint within $k$ days, would either result in drastic NPIs later or even commit to future, unavoidable violations of the constraint. 
In practice, the short-term forecast represents a hard cutoff on response times, for example, to mobilize additional ICUs and provides a guaranteed (within the accuracy of the model) window of time for interventions.

\begin{figure*}[t]
    \centering
    \includegraphics[width = 0.99\linewidth]{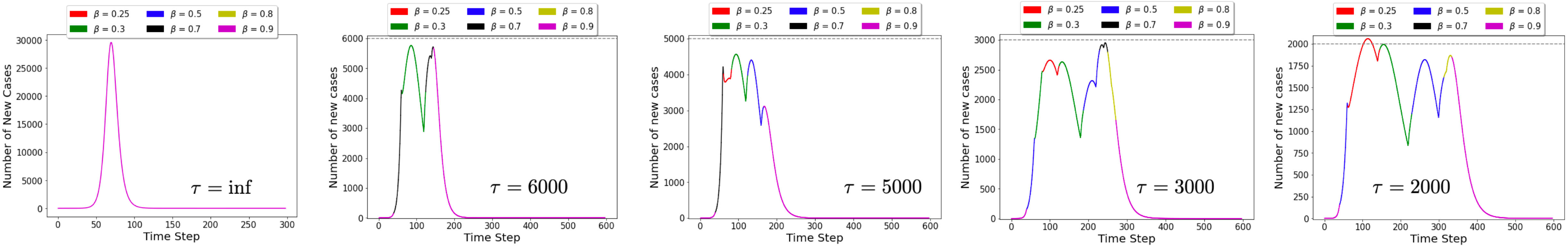}
    \caption{$\beta$ Optimization for SEIR - Policies inferred using the proposed approach with an SEIR model. The NPI choices comprised of different settings for the infectious rate $\beta$, which is known to be strongly linked to interventions. We show the policies obtained for different values of the threshold $\tau$.}
    \label{fig:seir-tau}
\end{figure*}

The long-term reward is evaluated for an additional $k_s$ days using a potentially different NPI. The search explores the NPI from the first $k$ days as well as all more stringent NPIs. However, unlike the short-term reward which is either given for all $k$ days or not at all, long-term rewards are awarded relative to the number of days the selected policy remains within the given constraint.
Therefore, a more relaxed policy that violates the constraint at some point can accumulate a higher reward than a more restrictive one that does not do so. 
Similar to the short-term reward, the limit to $k_s$ days reduces the total simulation time necessary for the optimization.
Furthermore, we reduce the number of explored scenarios by only considering more restrictive NPIs. While this does artificially limit the total predicted reward, it is unlikely to affect the final policy choice. Assuming a more relaxed NPI is acceptable past the initial $k$ days, it will always be explored in the next outer loop in $d < k$ days. 

Conceptually, the long-term reward reflects an optimistic choice to relax NPIs for the next $d$ days even if they are predicted to become problematic within $k+k_s$ days as long as tightening restrictions after $k$ days can correct for any violations.
An additional benefit of providing users the ability to choose these time scales is that it balances the level of risk with the uncertainties inherent in forecasting and enables one to compensate for external factors.
For example, if we expect new treatments to become available choosing a smaller $k_s$ results in more aggressive policy choices while still considering the need to correct for problems if this hope does not materialize.
A detailed description is given by Algorithm~\ref{alg}. Given an \e~model $\mathrm{E}$, its initial state $\mathrm{X}[0]$ and the threshold $\tau$ as inputs, the policy optimization assumes that NPIs can be changed every $d$ time steps.
There is no additional penalty to perform the switch which could easily be included if necessary.

\paragraph{Surrogate Model.}At the core of our approach is the need to obtain look-ahead estimates of the pandemic state given the current state $\mathrm{X}[t]$ and the NPI $n[t]$, which in turn requires evaluation of the \e~model $\mathrm{E}$. In cases where this evaluation is computationally expensive, it is beneficial to build a machine learned surrogate model that predicts the future states given the current state and the NPI choice. Formally, we build the surrogate to produce $k-$step predictions, $\mathrm{X}[t+1:t+K] = \hat{F}(X[t]; n[t])$, where $k$ is the look-ahead parameter. Note that, since the ODE-solver for SEIR is highly efficient, we do not use a surrogate in that case. However, the agent-based EpiCast model is computationally expensive and hence we build a machine learning surrogate for faster evaluation. The details of the surrogate model are provided in Section \ref{sec:epi}.

\begin{figure}[t]
    \centering
    \includegraphics[width = 0.5\linewidth]{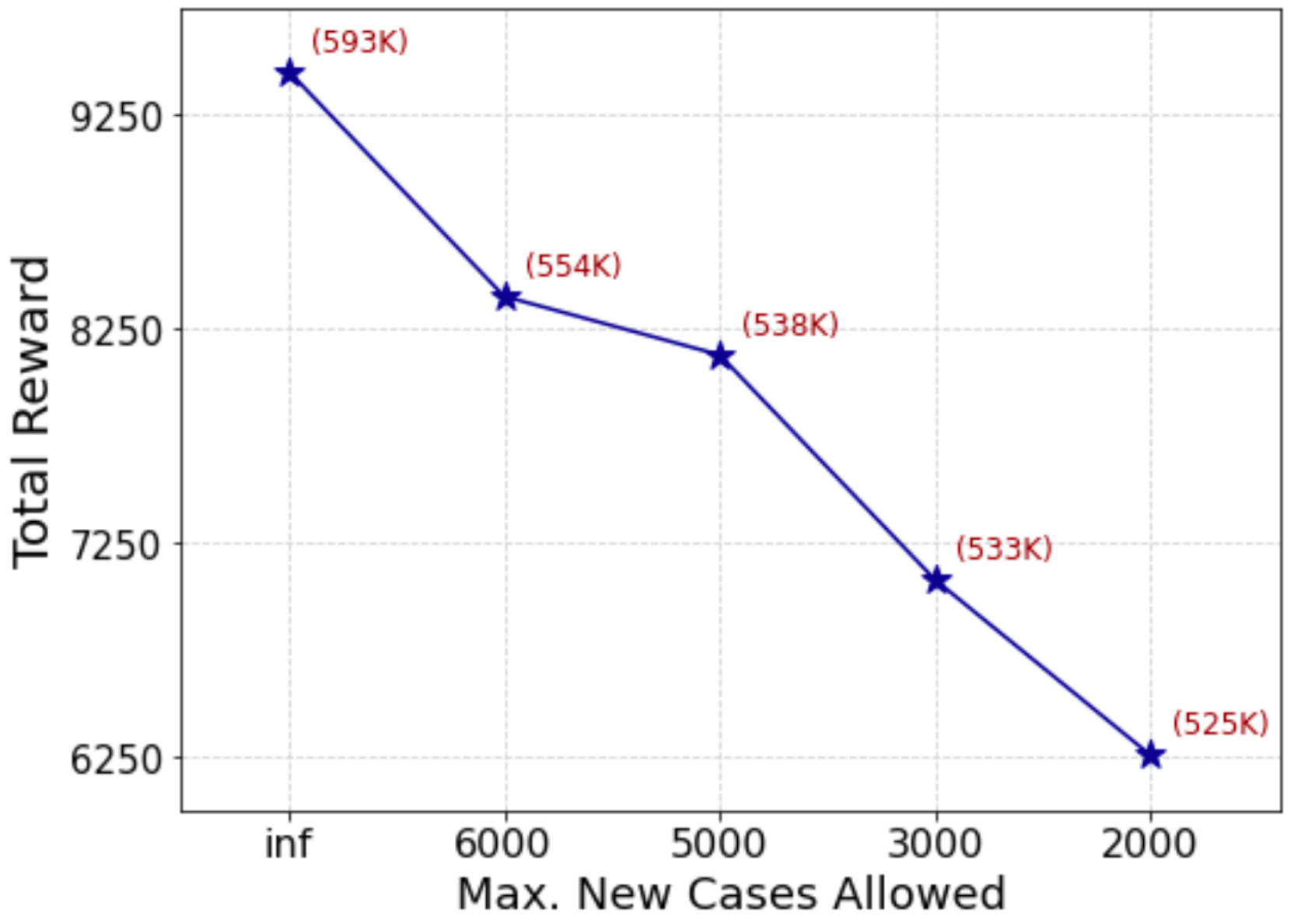}
    \caption{Effect of $\tau$ on the optimal policy. For each case, we show the total reward and the cumulative number of infections (in parentheses). We find that as $\tau$ becomes lower, we can obtain more conservative policies in terms of infections, but comes at the price of diminished rewards.}
    \label{fig:seir-reward}
\end{figure}


\section{Case Studies}
In this section, we use the proposed optimization algorithm to determine the NPI policy using both SEIR and EpiCast models. All our empirical studies are carried out using different scenarios, and we employ a simple reward assignment for each of the NPI choices. In both the models, we set the frequency of changing the NPI $d = 14$ days. As described earlier, while SEIR reflects the effects of NPI coarsely through the change in infectious rate $\beta$, EpiCast allows more fine-grained characterization of school or business restrictions.

\subsection{Optimizing $\beta$ Switches in SEIR}
In the case of SEIR, we consider the scenario where we are at the beginning of a pandemic, i.e $I(0) = 0$, $E(0) = 1$, $R(0) = 0$. Since the four compartments in the SEIR model add up to the total population, the susceptible compartment $S(0) = N - 1$. In this study, the population $N$ is set to $3e6$. As discussed earlier, following existing work on COVID-19, we set the incubation rate $\sigma = 0.2$ and recovery rate $\gamma = 0.1$. Given the set of NPI choices in the form of $\beta$ values (infectious rate) $\mathcal{N}= \{0.25,0.3,0.5,0.7,0.8,0.9\}$ and the corresponding reward assignments $\mathcal{C} \coloneqq \{1, 3, 6, 8, 12, 15\}$. The higher the infectious rate $\beta$, more severe the infection is or less restrictive the interventions are. 

The na\"ive policy of using the most restrictive NPI will provide the highest reward, when there is no constraint on the limit on the number of new cases each day, i.e., the threshold $\tau = inf$. As showed in Figure \ref{fig:seir-tau}, this na\"ive policy peaks within the first 75 days and the number of new cases reaches as high as $30K$. This unrestricted policy can naturally lead to significant overheads to the healthcare system. To circumvent this, we perform the optimization in Algorithm 1, by placing an upper limit on the number of cases, $\tau = \{6000, 5000, 3000, 2000\}$. As one might expect, the policy selection over the entire period is highly non-trivial, given the combinatorial nature of this optimization process. In our algorithm, we set the look-ahead parameters $k = 21$ days and $k_s = 35$ days, i.e., a total of $8$ weeks. 

In Figure~\ref{fig:seir-tau}, we illustrate the policies inferred using our approach for different values of $\tau$. Interestingly, in all cases, our greedy optimization produces effective policies that meet the constraint. As one might expect, the policies become more complex as $\tau$ becomes lower. We make two key observations from the results. First, due to the constraint $\tau$, the epidemic takes a much longer duration to flatten; for example at $\tau=3000$ it takes about $350$ time steps when compared to $\sim 100$ time steps in the case of $\tau = inf$. Second, our optimization balances short-term and long-term rewards, thus producing policies that are not overly conservative. For example, unless the constraint $\tau$ becomes very strong (say $2000$), the most restrictive NPI ($\beta=0.25$) is rarely  necessary. This effectively balances between the exponential growth of the epidemic and societal impacts of total inactivity through highly restrictive interventions. 

\begin{figure*}[!t]
    \centering
    \includegraphics[width = 0.99\linewidth]{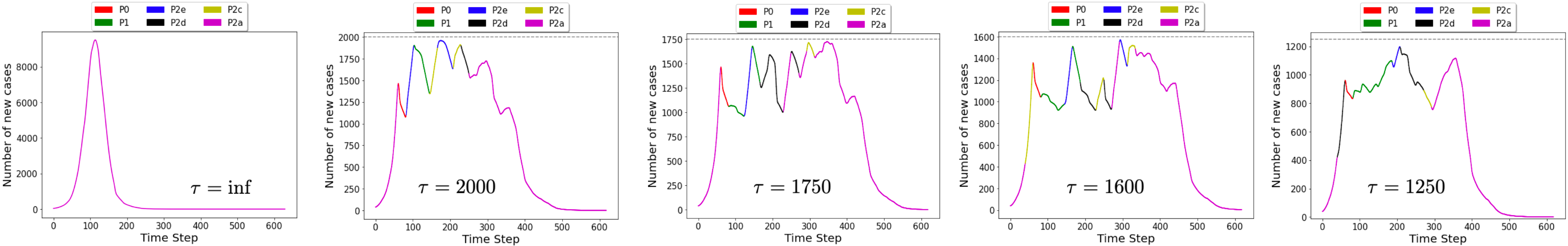}
    \caption{School opening policy optimization - Policies inferred using our approach with the EpiCast model. The NPI choices comprised of different phases of school opening. We show the policies obtained for varying values of the threshold $\tau$.}
    \label{fig:epi-school-tau}
\end{figure*}

\begin{figure*}[!t]
    \centering
    \includegraphics[width = 0.99\linewidth]{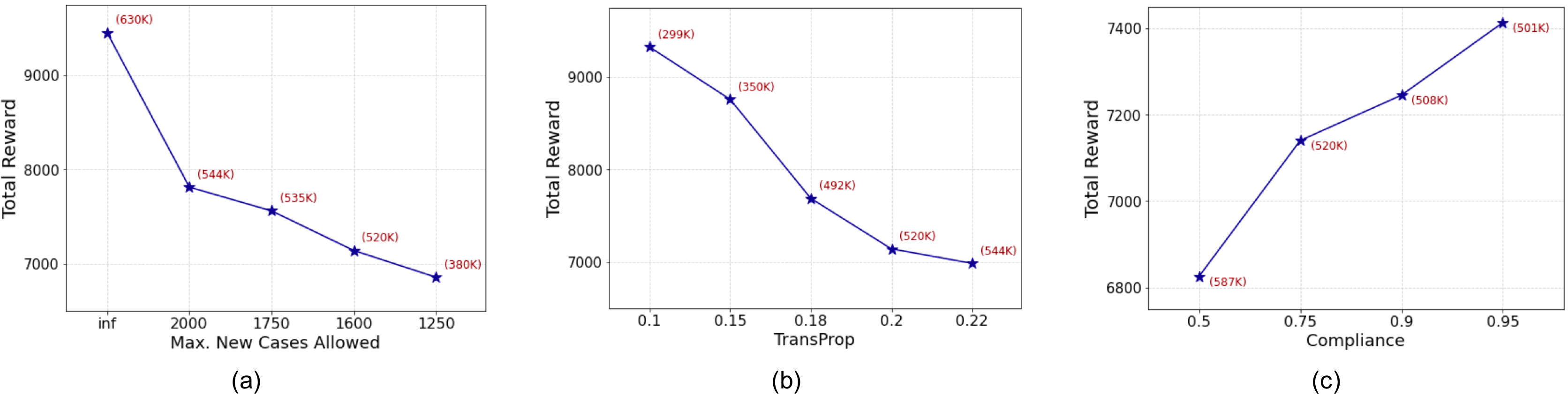}
    \caption{(a) Effect of $\tau$ on the cumulative infections and achieveable reward; (b) The impact of the transmission probability parameter on the policy. We find that higher TransProp leads to a spike in the total infections, while also reducing the reward significantly; (c) For a given the TransProp, the Compliance parameter reveals a positive impact on the quality of the policy.}
    \label{fig:epi-school-reward}
\end{figure*}

Since the duration for flattening is much longer with our policies, due to the use of less restrictive NPIs or no intervention ($\beta = 0.9$), it is important to study the total number of cases across the entire duration. This will indicate if the relaxations recommended by the inferred policy eventually leads to more (cumulative) cases. To study this trade-off,  Figure~\ref{fig:seir-reward} shows the total reward and the cumulative number of cases achieved using the different policies. Interestingly, with only a little compromise in the total reward, our policy achieves significant reduction in the total number of cases (indicated in parentheses near each marker). For example, using $\tau = 6000$ results in $\sim40$K lesser cases in total, when compared to $\tau=inf$. However, as the constraint on $\tau$ becomes more severe, i.e. $3000$ or $2000$, we see effects of herd immunity as sufficient number of cases still appear.

\subsection{Optimizing Phase Switches in EpiCast}
\label{sec:epi}
In this section, we describe how our optimization approach can be used with the agent-based EpiCast model.
When compared to the SEIR model, EpiCast can model the impact of fine-grained interventions (e.g., 2 days of school closure every week vs full closure) and hence can contain more complex dynamics, particularly when there is an interplay between different intervention choices.
This improved modeling comes at the price of high computational complexity as well as practical challenges in adjusting policies on the fly.
To make the problem tractable epidemiologist have chosen to split the different parameter choices into a set of discrete policies, i.e.\ a given school or work schedule, and a number of continuous parameters.
Furthermore, as these simulations require complex workflows and large scale parallel computing it is practically infeasible to change either policy or parameters during an individual run thus leaving parameter matching as well as policy optimization to an outer loop solution.
Here we explore two sets of policies on schools and workplaces within a six dimensional parameter space.
The six parameters are comprised of three global disease properties, namely the probability of transmission between individuals, the percentage of infections that are asymptomatic, and the so called relative infectiousness, and three parameters describing the current state of the epidemic: the number of currently infected individuals, the number of recovered (and assumed immune) patients, and the level of compliance to non-pharmaceutical interventions like masks.
The different policies roughly match the overall phases outlined in~\cite{OpeningUp} further separated by different schedules. 
More specifically, we explore Phase 0, with only essential businesses open and no schools (P$0$), Phase 1 which corresponds to slightly more businesses and less stringent distancing measures but with schools still universally closed (P$1$), and Phase 2 with even less restrictions on businesses and schools potentially opening. 
For schools our collaborators are interested in understanding the impact of different proposed schedules for in person education: 5 days of school (P$2$a), 3 days of school (P$2$c), 2 split cohorts each with 2 days of school (P$2$d); and 1 day of school (P$2$e). 
Similarly, for the industry scenarios we further separate the basic phases into different workplace schedules for all businesses open in the respective phase: a 5 day workweek (P$1$q, P$2$q); a 3 day work week (P$1$r, P$1$r); two split cohorts working 5 out of 10 days (P$1$v, P$2$v). Note that, in phase $1$ we assume that only a subset of the businesses to be open, while in phase $2$ the restrictions are relaxed and more businesses are operational. Note that, in practice it is unlikely that all workplaces would be able to implement all schedules for various reasons nor would all schools consider every variation and thus none of our results should be considered as recommendations. 
Instead, the goal is to develop a framework for policy makers that given a set of (potentially complex) scenarios expressed through \epi{} would provide suggestions on optimal policy choices given certain constraints. 
Since the computational costs make a direct policy optimization as in the SEIR model impossible we propose to first build surrogate models for the different policies each able to emulate \epi{} across a wide range of parameter settings.

\begin{figure*}[!t]
    \centering
    \includegraphics[width = 0.9\linewidth]{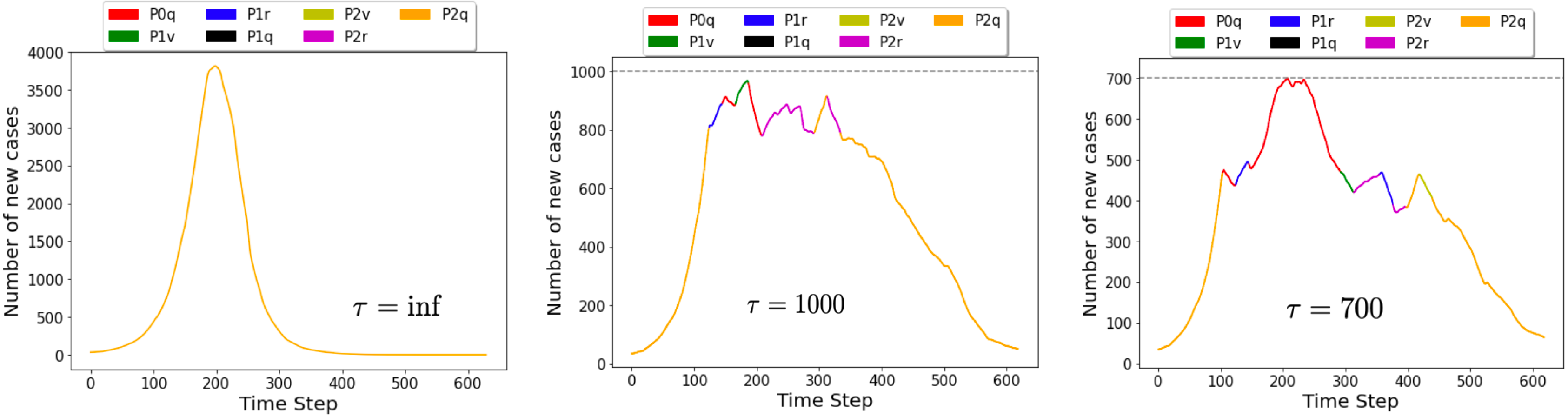}
    \caption{Business opening policy optimization - Policies inferred using our approach with the EpiCast model. The NPI choices comprised of different schedules for business opening. We show the policies obtained for varying values of the threshold $\tau$}
    \label{fig:epi-industry-tau}
\end{figure*}

\paragraph{Experiment Design for \epi{}.} While building surrogate models is typical in scientific problems~\cite{koziel2013surrogate}, the quality of the surrogates relies directly on the experiment design used to generate the dataset. To create the necessary training data for these models, we introduce an iterative approach aimed at reducing the computational costs as well as improving model fits.
The challenge in fitting a surrogate models to all phases of the disease is that large portions of the parameter space are invalid, in particular with respect to the initially infected and recovered populations, i.e.\ a large number of currently infected without anyone recovered.
While it is possible to execute EpiCast with any parameter setting, using unrealistic combinations wastes compute resources and likely affects the quality of the surrogate.
However, it is unclear \textit{a} priori which parameter configurations may be valid or how to sample from this space.
Instead, we first create simulation ensembles resembling realistic early outbreaks, i.e. few infected and recovered individuals and simulate the corresponding disease progression until outbreak has passed (typically 360 days or longer).
We then analyze all intermediate epidemic states and create additional simulations using these as starting conditions while varying the remaining disease parameters.
In our experience two iterations of this process, meaning three sets of simulation ensembles, are sufficient to cover the parameter space densely enough for surrogate modeling.

\paragraph{Surrogate Model.}Given the dataset generated using the experiment design described earlier, we build a machine-learned surrogate that can map the initial pandemic state $\mathrm{X}[0]$ to predict the future trajectory of the states. Since predicting long-term trends can be challenging with EpiCast, we resort to estimating the states for the next $21$ time steps. We create this dataset using sliding windows on the EpiCast samples and design a surrogate $\mathrm{F}$ that takes as input the current state ($6$-dimensional input) and outputs the trajectory curves for number of infected and removed cases. Following the recent surrogate modeling literature~\cite{anirudh2020improved}, we first constructed a low-dimensional latent space to better capture the structure in the curves. We explored the use of a simplified formulation with principal component analysis (PCA) on the curves (concatenated) as well as a more sophisticated multi-variate sequence-to-sequence models (autoencoders). We find that both these strategies are capable of accurately representing the short-term dynamics, as measured using reconstruction error of held-out validation data. Note that, we constructed a latent space of $5$ dimensions for each of the NPI choices considered (different school closing schedules). This pre-training step reformulates the surrogate modeling problem as predicting into the latent space, in lieu of the curves directly. The surrogate model was implemented as a fully connected network with $5$ hidden layers (configuration $[64,128,256,64,32]$) and ReLU activations. We trained the model with the mean-squared error objective and an $\ell_2$ regularizer on the weights using the Adam optimizer with learning rate $0.001$. We also compared the performance of this model against a random forests regressor containing $100$ trees and found the fully connected network to be marginally better -- $R$-squared statistics of $0.94$ and $0.93$ respectively.

\paragraph{School Closure Policy Optimization.}In this study, we considered a scenario where the total population for a region of interest was set to $1e7$ and the initial state of the pandemic was set to the following values:
\begin{itemize}
    \item Number of infected: $250$; Number of removed: $25$K
    \item TransProp: $0.2$; Asymptomatic ratio: $0.3$
    \item Relative infectiousness: $0.9$; Compliance: $0.75$
\end{itemize}The set of NPI choices comprised of $6$ different settings under phases 0,1 and 2 ($\mathcal{N}$ = [P$0$,P$1$,P$2$e, P$2$d, P$2$c, P$2$a]) with the corresponding rewards $\mathcal{C} = [1,3,6,8,12,15]$. Similar to the previous experiment, we set the frequency of NPI change at $d = 14$ days and used the look-ahead parameters $k = 21$ and $k_s = 35$ days respectively. We ran the proposed algorithm with the pre-trained surrogates for each of the $6$ NPI settings and computed the optimal mitigation policy for a duration of $600$ time-steps. The most relaxed policy P$2$a corresponds to the case of $\tau=inf$ and achieves the maximum reward. However, as showed in Figure \ref{fig:epi-school-tau}, one can obtain more realistic policies by placing a constraint on the number of new cases. In particular, as we reduce $tau$ from $2000$ to $1250$ we observe that the policy rolls back to a very restricted setting (P$0$ or P$1$) for longer periods of time, indicating that the exponential nature of the epidemic requires some rather stringent constraints especially early on to get to a more controlled state of the epidemic. Furthermore, all results suggest that the primary effect of the restrictions is to broaden the peak until the epidemic has run its course and even the most relaxed policy shows an steep decline in cases. Interestingly, as illustrated in Figure~\ref{fig:epi-school-reward}(a), the optimal policy at $\tau = 2000$ achieves a significant reduction in the total number of infections ($544$K) when compared to the $\tau=inf$ case ($630$K) for $\approx 15\%$ drop in the total reward. This clearly shows the efficacy of our look-ahead optimization in inferring non-trivial policies, while taking into account the complex interplay between the different NPI choices. 

\noindent \textit{Impact of TransProp} It is well-known to epidemiologists that the probability of transmission between individuals is a critical parameter in controlling the trajectory of infections. Hence, we studied the impact of this parameter on the policies inferred using our algorithm. For this purpose, we varied the TransProp parameter between $0.1$ and $0.22$ while fixing the rest at the settings specified earlier. Note that, for this analysis ,we fixed the threshold $\tau = 1600$. From the result in Figure~\ref{fig:epi-school-reward}(b), it is apparent that the efficacy of the policy becomes increasingly inferior as TransProp grows -- increase in the total number of cases as well as a significant reduction in the total reward.

\noindent \textit{Effect of Compliance} Another important aspect of mitigating epidemics such as COVID-19 is the public compliance to the mandates enforced and EpiCast includes that as one of its input parameters. We studied its effect on the policies inferred by varying compliance between $0.5$ and $0.95$ (higher value implies more compliant), while fixing the rest of the parameters as specified earlier. Similar to the previous experiment, we fixed $\tau = 1600$. As showed in Figure~\ref{fig:epi-school-reward}(c), improved compliance does lead to better policies both in reduced total number of infections and increased reward. However, given the relatively high transmission probability of $0.2$, there is a limit beyond which the reward could not be increased even with a high compliance.

\paragraph{Industry Closure Policy Optimization.} In this experiment, we considered NPI choices pertinent to business opening and followed the protocol from the school opening study for running the policy optimization. Given $\mathcal{N}=$[P$0$,P$1$v,P$1$r,P$1$q,P$2$v,P$2$r,P$2$q] and the corresponding reward assignments $\mathcal{C} = [1,3,6,8,12,15,18]$, we obtained the policies in Figure~\ref{fig:epi-industry-tau}. First, we find that for the same initial settings as before, the trajectory for the infection growth is less severe when only businesses are open (schools closed) when compared to the case where schools are also open. Even when we aggressively reduced the threshold $\tau$ to $700$, we could find an optimal policy, which however required a brief period of returning to phase $0$ (total closure).


\section{Conclusions}
In this paper, we proposed an epidemic mitigation policy optimization algorithm that systematically accounts for short- and long-term effects of non-pharmaceutical interventions, via sophisticated \e~models. Our approach provides a principled, yet scalable, way to navigate through the combinatorial choices of interventions. Using case studies with SEIR and agent-based EpiCast models, we demonstrated the efficacy of our greedy algorithm in determining non-trivial policies (high reward) that satisfy the constraint on the number of infections. This can be used by policy makers and epidemiologists to gain critical insights into the interplay between different NPIs and current state of the epidemic.

 \section{Acknowledgements}
 This work was performed under the auspices of the U.S. Department of Energy by
 Lawrence Livermore National Laboratory under Contract DE-AC52-07NA27344. Research was supported by the DOE Office of Science through the National Virtual Biotechnology Laboratory, a consortium of DOE national laboratories focused on response to COVID-19, with funding provided by the Coronavirus CARES Act.

\bibliography{refs}

\end{document}